\documentclass[11pt,letterpaper]{article}
\usepackage{emnlp2017}
\usepackage{times}
\usepackage{latexsym}
\usepackage{CJK,algorithm,algorithmic,amssymb,amsmath,array,epsfig,graphics,enumerate,multirow,array,float,subfigure,verbatim,epstopdf}
\usepackage{times}
\usepackage{latexsym}
\usepackage{enumitem}
\usepackage{url} %2
\usepackage{color}
\usepackage{hhline}

\emnlpfinalcopy

\newcommand{\vpara}[1]{\vspace{0.05in}\noindent\textbf{#1 }}

\definecolor{new1}{rgb}{0.8,0.4,0}

\title{Zero-Shot Transfer Learning for Event Extraction}
\author{Lifu Huang \and Heng Ji \\ Rensselaer Polytechnic Institute \\ \{huangl7, jih\}@rpi.edu
        \AND
        Kyunghyun Cho \\ New York University \\  kyunghyun.cho@nyu.edu \And
        Clare R. Voss \\ US Army Research Lab \\  clare.r.voss.civ@mail.mil }
\date{}

\begin{document}

\maketitle
%\lifu{Remove Draft Command before Submission}

\begin{abstract}

%\clare{ 1) the paper should pivot its focus just a bit to highlight that there are TWO pre-existing resources being leveraged:
% the small annotated corpus with ``seen'' types AND 
% the event ontologies with ``seen'' and ``unseen'' types}

%\clare{1) should help head off critiques like ACL reviewer 1 who wanted to see a comparison to Lampert et al, -  they were failing to see that this paper's approach DOES have an analogue to L et al's attribute table, namely a combination of (i) the ontologies as external resource injected into (ii) the constructed shared semantic space, that then yields transferable information.} 

%\clare{ 2)  if the title has zero-shot in it, then the abstract ought to connect to that concept in its wording somewhere - not sure how best to do that though}

%\clare{3) I added italics for the featured (best?) result at the end to emphasize this is the punchline but the italics are not strictly needed 
% }

%limiting their performance on new (``unseen'') 

Most previous event extraction studies have relied heavily on features derived from annotated event mentions, thus cannot be applied to new event types without annotation effort. In this work, we take a fresh look at event extraction and model it as a grounding problem. We design a transferable neural architecture, mapping event mentions and types jointly into a shared semantic space using structural and compositional neural networks, where the type of each event mention can be determined by the closest of all candidate types %\heng{need to revise this description if we end up with using GraphCNN}
. By leveraging (1)~available manual annotations for a small set of existing event types and (2)~existing event ontologies, our framework applies to new event types without requiring additional annotation. Experiments on both existing event types (e.g., ACE, ERE) and new event types (e.g., FrameNet) demonstrate the effectiveness of our approach. \textit{Without any manual annotations} for 23 new event types, our zero-shot framework achieved performance comparable to a state-of-the-art supervised model which is trained from the annotations of 500 event mentions.

\end{abstract}

% We are developing new techniques to transfer knowledge from existing old event types to new event schema discovery, and transfer knowledge from high-resource languages to low-resource languages for event extraction. Experiments on Chinese and Spanish event nugget detection are promising.

%\heng{consider to use the figure in DEFT event talk and put it here. The examples need to be made clearer.}

%In our previous work (Huang et al., 2016), we developed a Liberal Information Extraction framework to discover event schema and extract events and arguments simultaneously. However, because many manually constructed event schemas exist for some common event types in certain domains, we will apply zero-shot transfer learning to transfer knowledge about existing event types to new event types, and from high-resource languages to low-resource languages. 

\section{Introduction}
\label{sec:intro}

\begin{figure*}[!htb]
\centering
\includegraphics[width=.93\textwidth]{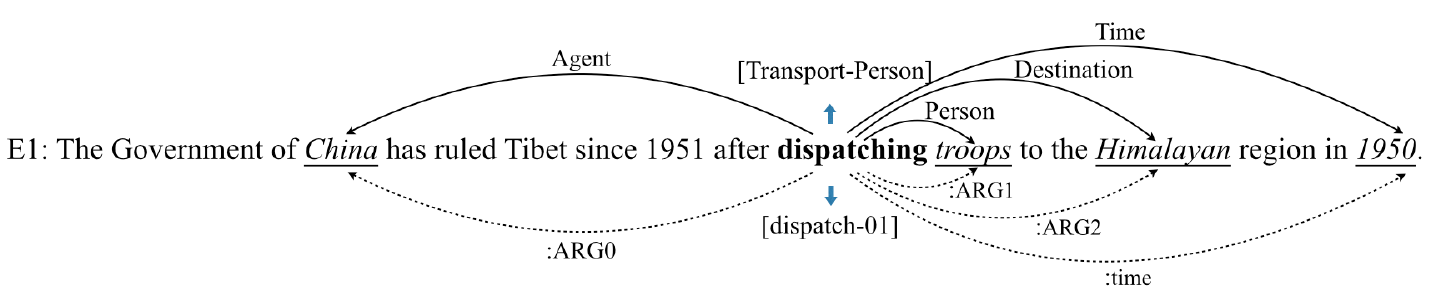}
\vspace{-0.3cm}
\caption{Event Mention Example: \textbf{dispatching} is the trigger of a \textit{Transport-Person} event with four arguments.}
\label{eventExample}
\vspace{-0.5cm}
\end{figure*}

The goal of event extraction is to extract event triggers and arguments from unstructured data. An example is shown in Figure~\ref{eventExample}. Major obstacles to making progress on event extraction have been the poor portability of traditional supervised methods and the limited coverage of available event annotations. Handling new event types means to start from scratch without being able to re-use annotations for old event types. The main reason is that these approaches modeled event extraction as a classification problem, encoding features only by measuring the similarity between rich features encoded for test event mentions and annotated event mentions. In these models, an event type (e.g., \textit{Transport-Person}) or an argument role (e.g., \textit{Destination}) is simply treated as an atomic symbol (i.e., a surface lexical form). Therefore it's not feasible to repeat the high-cost annotation process for each of the 3,000+ event types.

In fact, many rich event ontologies have been recently developed, including FrameNet~\cite{Baker2003}, VerbNet~\cite{Kipper2008}, Propbank~\cite{Palmer2005} and OntoNotes~\cite{pradhan2007ontonotes}, where each event type is associated with a set of predefined argument roles. We observe that both event mentions and types can be represented with structures, where event mention structure is constructed from trigger and candidate arguments, and event type structure consists of event type and predefined roles. Consider two example sentences:

\begin{enumerate}[label={E}\arabic*.]\itemsep0pt
\fontsize{10.5pt}{13pt}\selectfont
\item The Government of \underline{\textit{China}} has ruled Tibet since 1951 after \textbf{dispatching} \underline{\textit{troops}} to the \underline{\textit{Himalayan}} region in \underline{\textit{1950}}. 
\item Iranian state television stated that the \textbf{conflict} between the \underline{\textit{Iranian police}} and the drug \underline{\textit{smugglers}} took place near the town of \underline{\textit{mirjaveh}}.
\end{enumerate}

\begin{figure}[h]
\centering
\includegraphics[width=.45\textwidth]{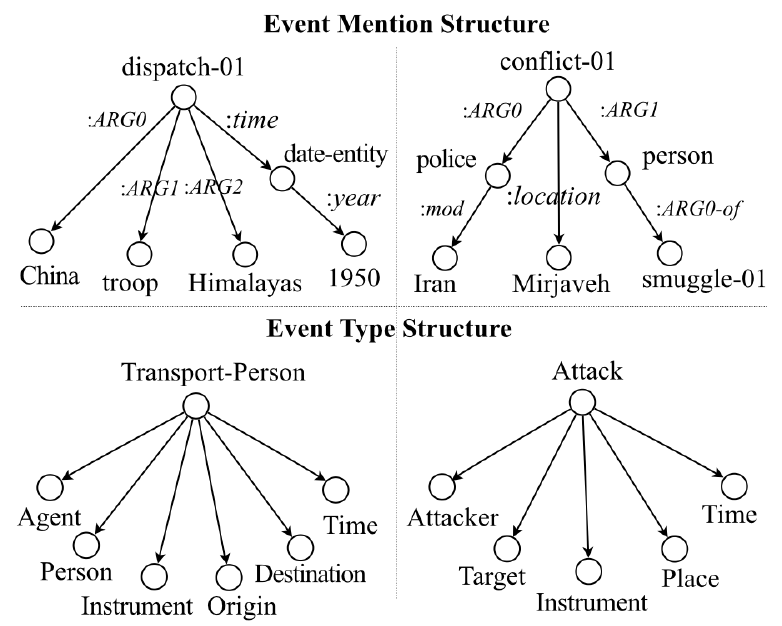}
\vspace{-0.2cm}
\caption{Examples of Event Mention and Type Structures from ERE.}
\label{structure}
\vspace{-0.2cm}
\end{figure}

E1, as can be seen in Figure~\ref{eventExample}, includes a \textit{Transport\_Person} event mention triggered by \textbf{dispatching} and E2 includes an \textit{Attack} event mention triggered by \textbf{conflict}. 
For each event mention, we apply Abstract Meaning Representation (AMR)~\cite{banarescu2013abstract} to identify candidate arguments and construct event mention structures. Meanwhile, the two event types can also be represented with structures from ERE (Entity Relation Event)~\cite{song2015light}, as shown in Figure~\ref{structure}. We can see that, besides the lexical semantics that relates a trigger to its type, their structures also tend to be similar: a \textit{Transport\_Person} event typically involves a \textit{Person} instead of an \textit{Artifact} as the patient, while an \textit{Attack} event involves a \textit{Person} or \textit{Location} as an \textit{Attacker}. This observation is similar to the theory that ``the semantics of an event structure can be generalized and mapped to event mention structures in a systematic and predictable way''~\cite{pustejovsky1991syntax}. Inspired by this theory, we take a fresh look at the event extraction task and model it as a ``\textbf{grounding}'' problem, by mapping each mention to its semantically closest event type in the ontology.

One possible implementation of this idea is Zero-Shot Learning (ZSL), which has been successfully exploited in visual object classification~\cite{frome2013devise,norouzi2013zero,socher2013zero}. The main idea of applying ZSL for vision tasks is to represent both images and type labels in a multi-dimensional vector space separately, and then learn a regression model to map from image semantic space to type label semantic space based on annotated images for \textit{seen} labels. This regression model can be further used to predict the \textit{unseen} labels of any given image. 

In this paper, we apply ZSL to event extraction. Given an event ontology, where each event type is defined with a rich structure (e.g., argument roles), we call the event types with annotated event mentions as \textit{seen} types, while those without annotations as \textit{unseen}. Our goal is to effectively transfer the knowledge of events from \textit{seen} types to \textit{unseen} types, so we can extract event mentions of any types defined in the ontology. We design a transferable neural architecture, which jointly learns and maps the structural representations of both event mentions and types into a shared semantic space by minimizing the distance between each event mention and its corresponding type. For event mentions with \textit{unseen} types, their structures will be projected into the same semantic space using the same framework and assigned types with top-ranked similarity values. 

There are two appealing advantages of this new view, which also manifest our contributions:

\begin{enumerate}[label=$\bullet$]\itemsep0pt
\fontsize{10.5pt}{13pt}\selectfont
\item This mapping/ranking function is ``universal'' and independent of event types, thus we can transfer resources from existing types to new types without any additional annotation effort; 

\item Many existing event ontologies cover a wider range of event types, which allow us to extend the scope of event extraction from several dozen types to thousands of types. 
\end{enumerate}

\section{Approach}
\label{sec:approach}

\subsection{Overview}
\label{subsec:overview}

\begin{figure*}[!htb]
\centering
\includegraphics[width=.85\textwidth]{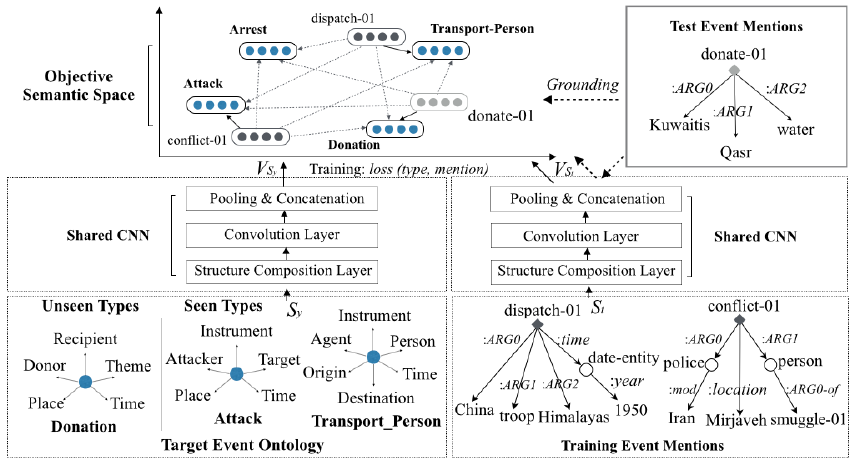}
\vspace{-0.3cm}
\caption{Architecture Overview}
\label{overview}
\vspace{-0.3cm}
\end{figure*}

Event extraction aims to extract both triggers and arguments. 
Figure~\ref{overview} illustrates the overall architecture of our approach for trigger typing, while argument typing follows the same pipeline.

Given a sentence $s$, we start by identifying candidate triggers and arguments based on AMR parsing~\cite{wang2015transition}. An example is shown in Figure~\ref{eventExample}. For each trigger $t$, e.g., \textit{dispatch-01}, we build a structure $S_t$ using AMR as shown in Figure~\ref{overview}. Each structure is composed of a set of tuples, e.g, $\langle$\textit{dispatch-01}, \textit{:ARG0}, \textit{China}$\rangle$. We use a matrix to represent each AMR relation, composing its semantics with two concepts for each tuple, and feed all tuple representations into CNN to generate event mention structure representation $V_{S_t}$. 

Given a target event ontology, for each type $y$, e.g., \textit{Transport\_Person}, we construct a type structure $S_y$ by incorporating its predefined roles, and use a tensor to denote the implicit relation between any types and arguments. We compose the semantics of type and argument role with the tensor for each tuple, e.g., $\langle$\textit{Transport\_Person}, \textit{Destination}$\rangle$. We generate the event type structure representation $V_{S_y}$ using the same CNN. By minimizing the semantic distance between \textit{dispatch-01} and \textit{Transport\_Person} using $V_{S_t}$ and $V_{S_y}$, we jointly map the representations of event mention and event types into a shared semantic space, where each mention is closest to its annotated type. 

After training, the compositional functions and CNNs can be further used to project any new event mention (e.g., \textit{donate-01}) into the semantic space and find its closest event type (e.g., \textit{Donation}).

\subsection{Candidate Trigger and Argument Identification}
\label{subsec:identification}

Similar to~\newcite{huangliberal}, we identify candidate triggers and arguments based on AMR Parsing~\cite{wang2015transition} and apply the same word sense disambiguation (WSD) tool~\cite{zhong2010makes} to disambiguate word senses and link each sense to OntoNotes, as shown in Figure~\ref{eventExample}. 

Given a sentence, we consider all noun and verb concepts that can be mapped to OntoNotes senses by WSD as candidate event triggers. In addition, the concepts that can be matched with verbs or nominal lexical units in FrameNet are also considered as candidate triggers. For each candidate trigger, its candidate arguments are specified by a subset of AMR relations, as shown in Table~\ref{AMRRelation}. 

\begin{table}[!htb]
\footnotesize
\begin{center}
\begin{tabular}{c|p{5cm}<{\centering}}
\hline
Categories  & Relations \\ \hline 
Core roles  & ARG0, ARG1, ARG2, ARG3, ARG4  \\ \hline
Non-core roles  & mod, location, instrument, poss, manner, topic, medium, prep-X\\ \hline
Temporal & year, duration, decade, weekday, time \\ \hline
Spatial & destination, path, location \\ \hline
\end{tabular}
\end{center}
\vspace{-0.2cm}
\caption{Event-Related AMR Relations.}
\label{AMRRelation}
\vspace{-0.3cm}
\end{table}

\subsection{Structure Construction and Composition}
\label{sec:structureComposition}

As Figure~\ref{overview} shows, for each candidate trigger $t$, we construct its event mention structure $S_t$ based on its candidate arguments and AMR parsing. For each type $y$ of the target event ontology, we construct a structure $S_y$ by incorporating its pre-defined roles and take the type as the root.

Each $S_t$ or $S_y$ is composed of a collection of tuples. For each event mention structure, a tuple consists of two AMR concepts and a AMR relation, while for each event type structure, a tuple consists of a type name and an argument role name. Here we propose two approaches to incorporate the semantics of relations into the two words of each tuple.

\vpara{Event Mention Structure} For each tuple $u = \langle w_1, \lambda, w_2\rangle$ in an event mention structure, we use a matrix to represent each AMR relation, and compose the semantics of the AMR relation $\lambda$ to the two concepts $w_1$ and $w_2$ as:
\vspace{-0.1cm}
\begin{displaymath}
V_{u} = [V_{w_1}^{'}; V_{w_2}^{'}] = f([V_{w_1}; V_{w_2}]\cdot M_{\lambda})
\vspace{-0.1cm}
\end{displaymath}
where $V_{w_1}$, $V_{w_2}$ $\in \mathbb{R}^{d}$ are the vector representations of words $w_1$ and $w_2$. 
$d$ is the dimension size of each word vector. $[\ ;\ ]$ denotes the concatenation of two vectors. $M_{\lambda}\in\mathbb{R}^{2d\times 2d}$ is the matrix representation for AMR relation $\lambda$. $V_{u}$ is the composition representation of tuple $u$, which consists of two updated vector representations $V_{w_1}^{'}$, $V_{w_2}^{'}$ for $w_1$ and $w_2$ by incorporating the semantics of $\lambda$.

\vpara{Event Type Structure} For each tuple $u^{'}=\langle y, r \rangle$ in an event type structure, where $y$ denotes the event type and $r$ denotes an argument role, following~\newcite{socher2013recursive}, we assume an implicit and ``universal'' relation between any pairs of type and argument, and use a single and powerful tensor to represent the implicit relation:

\begin{displaymath}
V_{u^{'}} = [V_{y}^{'}; V_{r}^{'}] = f([V_{y}; V_{r}]^{T}\cdot U^{[1:2d]}\cdot [V_{y}; V_{r}])
\end{displaymath}
where $V_{y}$ and $V_{r}$ are vector representations for $y$ and $r$. $U^{[1:2d]}\in \mathbb{R}^{2d\times 2d \times 2d}$ is a 3-order tensor. $V_{u}^{'}$ is the composition representation of tuple $u^{'}$, which consists of two updated vector representations $V_{y}^{'}$, $V_{r}^{'}$ for $y$ and $r$ by incorporating the semantics of their implicit relation $U^{[1:2d]}$.

\subsection{Joint Event Mention and Type Label Embedding}
\label{subsec:trigger}

CNN is good at capturing sentence level information in various Natural Language Processing tasks. In this work, we use it to generate structure-level representations. For each event mention structure $S_{t}=(u_1, u_2, ..., u_h)$ and each event type structure $S_{y}=(u^{'}_1, u^{'}_2, ..., u^{'}_p)$, which contains $h$ and $p$ tuples respectively, we apply a weight-sharing CNN to each input structure to jointly learn event mention and type structural representations, which will be later used to learn the ranking function for zero-shot event extraction.

\vpara{Input layer} is a sequence of tuples, where the order of tuples is from top to bottom in the structure. Each tuple is represented by a $d\times 2$ dimensional vector, thus each mention structure and each type structure are represented as a feature map of dimensionality $d\times 2h^{*}$ and $d\times 2p^{*}$ respectively, where $h^{*}$ and $p^{*}$ are the maximal number of tuples for event mention and type structures. We use zero-padding to the right to make the volume of all input structures consistent.

\vpara{Convolution layer}Take $S_t$ with $h^{*}$ tuples: $u_1, u_2, ..., u_{h^{*}}$ as an example. The input matrix of $S_t$ is a feature map of dimensionality $d\times 2h^{*}$.  We make $c_i$ as the concatenated embeddings of $n$ continuous columns from the feature map, where $n$ is the filter width and $0<i<2h^{*}+n$. A convolution operation involves a filter $W\in \mathbb{R}^{nd}$, which is applied to each sliding window $c_i$:
\vspace{-0.1cm}
\begin{displaymath}
\footnotesize
c^{'}_{i} = \tanh(W\cdot c_i + b)
\vspace{-0.1cm}
\end{displaymath}
where $c^{'}_{i}$ is the new feature representation, and $b\in \mathbb{R}^{d}$ is a biased vector. We set filter width as 2 and stride as 2 to make the convolution function operate on each tuple with two input columns.

\paragraph{Max-Pooling: } All tuple representations $c^{'}_{i}$ are used to generate the representation of the input sequence by max-pooling.

\paragraph{Learning: }
For each event mention $t$, we name the correct type as \textit{positive} and all the other types in the target event ontology as \textit{negative}. To train the composition functions and CNN, we first consider the following hinge ranking loss:

\vspace{-0.3cm}
\begin{displaymath}
\footnotesize
L_{1}(t, y)= \sum_{j\in Y, \ j\neq y} \max\{0, m-C_{t,y} + C_{t, j} \}
\end{displaymath}
\begin{displaymath}
\footnotesize
C_{t,y} = \cos([V_{t};V_{S_t}], [V_{y};V_{S_y}])
\end{displaymath}
where $y$ is the positive event type for $t$. $Y$ is the type set of the event ontology. $[V_{t};V_{S_t}]$ denotes the concatenation of representations of $t$ and $S_t$. $j$ is a negative event type for $t$ from $Y$. $m$ is a margin. 
$C_{t,y}$ denotes the cosine similarity between $t$ and $y$.

The hinge loss is commonly used in zero-shot visual object classification task. However, it tends to overfit the seen types in our experiments. While clever data augmentation can help alleviate overfitting, we propose two strategies: (1) we add ``negative'' event mentions into the training process. Here a ``negative'' event mention means that the mention has no positive event type among all seen types, namely it belongs to \textit{Other}. (2) we design a new loss function as follows: 
\begin{displaymath}
\scriptsize
L^{d}_{1}(t, y)=
\begin{cases}
\max\limits_{j\in Y, j\neq y} \max\{0, m-C_{t, y} + C_{t, j} \}, \quad\quad y\neq Other \\ 
\max\limits_{j\in Y^{'}, j\neq y^{'}} \max\{0, m-C_{t, y^{'}} + C_{t, j} \}, \quad y=Other
\end{cases}
\end{displaymath}
where $Y$ is the type set of the event ontology. $Y^{'}$ is the seen type set. $y$ is the annotated type. $y^{'}$ is the type which ranks the highest among all event types for event mention $t$, while $t$ belongs to \textit{Other}. 

By minimizing $L^{d}_{1}$, we can learn the optimized model which can compose structure representations and map both event mention and types into a shared semantic space, where the positive type ranks the highest for each mention. 

\subsection{Joint Event Argument and Role Embedding}
\label{subsec:argument}

For each mention, we map each candidate argument to a specific role based on the semantic similarity of the argument path. Take E1 as an example. \textit{China} is matched to \textit{Agent} based on the semantic similarity between \textit{dispatch-01}$\rightarrow$ \textit{:ARG0}$ \rightarrow$ \textit{China} and \textit{Transport-Person}$\rightarrow$\textit{Agent}. 

Given a trigger $t$ and a candidate argument $a$, we first extract a path $S_a = (u_1, u_2, ..., u_p)$, which connects $t$ and $a$ and consists of $p$ tuples. Each predefined role $r$ is also represented as a structure by incorporating the trigger type, $S_r = \langle y, r\rangle$. We apply the same framework to take the sequence of tuples contained in $S_a$ and $S_r$ into a weight-sharing CNN to rank all possible roles for $a$. 
\begin{displaymath}
\scriptsize
L^{d}_{2}(a, r)= 
\begin{cases}
\max\limits_{j\in R_{y}, j\neq r} \max\{0, m-C_{a, r} + C_{a, j} \} \quad \quad r\neq Other \\ 
\max\limits_{j\in R_{Y^{'}}, j\neq r^{'}} \max\{0, m-C_{a, r^{'}} + C_{a, j} \} \ \ r|y=Other
\end{cases}
\end{displaymath}
where $R_{y}$ and $R_{Y^{'}}$ are the set of argument roles which are predefined for trigger type $y$ and all seen types $Y^{'}$. $r$ is the annotated role and $r^{'}$ is the argument role which ranks the highest for $a$ when $a$ or $y$ is annotated as \textit{Other}.

In our experiments, we sample various size of ``negative'' training data for trigger and argument labeling respectively. In Section~\ref{subsec:ace} we describe how the negative training instances are generated. We adopt a pipelined framework and train the model for trigger labeling and argument labeling separately.

\subsection{Zero-Shot Classification}
\label{subsec:classification}
 
During test, given a new event mention $t^{'}$, we compute its mention structure representation for $S_{t^{'}}$ and all event type structure representations for $S_{Y} = \{S_{y_{1}}, S_{y_{2}}, ..., S_{y_{n}}\}$ using the same parameters trained from seen types. We rank all event types based on their similarity scores with mention $t^{'}$. The top ranked prediction for $t^{'}$ from the event type set, denoted as $\widehat{y}(t^{'}, 1)$, is given by:
\vspace{-0.03cm}
\begin{displaymath}
\footnotesize
\widehat{y}(t^{'}, 1) = \arg\max_{y\in Y}{ \cos([V_{t^{'}};V_{S_{t^{'}}}], [V_{y}; V_{S_{y}}])}
\vspace{-0.2cm}
\end{displaymath}

Moreover, $\widehat{y}(t^{'}, k)$ denotes the $k^{th}$ most probable event type predicted for $t^{'}$. We will investigate the event extraction performance based on the top-$k$ predicted event types.

After determining the type $y^{'}$ for mention $t^{'}$, for each candidate argument, we adopt the same ranking function to find the most appropriate role from the role set defined for $y^{'}$. 
 
\section{Experiments}
\label{sec:exp}

\subsection{Hyper-Parameters}
\label{subsec:data}

We use an August 11, 2014 English Wikipedia dump to learn trigger sense and argument embeddings based on the Continuous Skip-gram model~\cite{mikolov2013efficient}. Table~\ref{parameters} shows the hyper-parameters we used to train models. 

\begin{table}[h]
\footnotesize
\begin{tabular}{p{6.0cm}<{\centering}|p{0.7cm}<{\centering}}
\hline 
Parameter Name & Value \\
\hline
Word Sense Embedding Size & 200 \\
Initial Learning Rate & 0.1 \\
\# of Filters in Convolution Layer & 500 \\
Maximal \# of Tuples for Mention Structure &  10 \\
Maximal \# of Tuples for Argument Path & 5 \\
Maximal \# of Tuples for Event Type Structure &  5 \\
Maximal \# of Tuples for Argument Role Path & 1 \\
\hline
\end{tabular}
\vspace{-0.1cm}
\caption{Hyper-parameters.}
\label{parameters}
\vspace{-0.35cm}
\end{table}

\subsection{ACE Event Classification}
\label{subsec:ace}
We first use ACE event schema as our target event ontology and assume the boundaries of triggers and arguments are given. Of the 33 ACE event types, we select the top-$N$ most popular event types from ACE05 data as ``seen'' types, and use 90\% event annotations of these for training and 10\% for development. $N$ is set as 1, 3, 5, 10 respectively. We test the zero-shot classification performances on the annotations for the remaining 23 unseen types. Table~\ref{seentypes} shows the types that we selected for training in each experiment setting.

The negative event mentions and arguments that belong to \textit{Other} are sampled from the output of the system developed by~\newcite{huangliberal} based on ACE05 training sentences, which groups all candidate triggers and arguments into clusters based on semantic representations and assigns a type/role name to each cluster. We sample the negative event mentions from the clusters (e.g., \textit{Build}, \textit{Threaten}) which cannot be mapped to ACE event types. We sample the negative arguments from the arguments associated with these negative event mentions. Table~\ref{aceStatistics} shows the statistics of the training, development and testing data sets. 
\begin{table}[!htp]
\footnotesize
\begin{tabular}{p{0.7cm}<{\centering}|p{0.32cm}<{\centering}|p{5.15cm}<{\centering}}
\hline 
Setting & N & Seen Types for Training/Dev  \\
 \hline
A & 1 & Attack \\
B & 3 & Attack, Transport, Die \\
C & 5 & Attack, Transport, Die, Meet, Arrest-Jail   \\
D & 10 & Attack, Transport, Die, Meet, Sentence, Arrest-Jail, Transfer-Money, Elect, Transfer-Ownership, End-Position \\
\hline
\end{tabular}
\vspace{-0.1cm}
\caption{Seen Types in Each Experiment Setting.}
\vspace{-0.2cm}
\label{seentypes}
\end{table}

\begin{table*}[!htp]
\footnotesize
\begin{tabular}{p{0.8cm}<{\centering}|p{1.4cm}<{\centering}|p{1.35cm}<{\centering}|p{1.4cm}<{\centering}|p{1.2cm}<{\centering}|p{1.4cm}<{\centering}||p{1.4cm}<{\centering}|p{1.2cm}<{\centering}|p{1.4cm}<{\centering}}
\hline 
\multirow{2}{0.8cm}{Setting Index} & \multicolumn{3}{p{3.8cm}<{\centering}|}{Training} & \multicolumn{2}{p{2.6cm}<{\centering}||}{Development} & \multicolumn{3}{p{3.8cm}<{\centering}}{Test}\\
\cline{2-9} 
& \# of Types/Roles & \# of Events & \# of Arguments & \# of Events & \# of Arguments & \# of Types/Roles & \# of Events & \# of Arguments \\
 \hline

A & 1/5 &   953/900  &  894/1,097 &  105/105 & 86/130  & \multirow{4}{0.8cm}{23/59}  & \multirow{4}{0.8cm}{753} & \multirow{4}{0.8cm}{879} \\
B & 3/14 &  1,803/1,500 & 2,035/1,791  &  200/200 & 191/237 &   &  &  \\
C & 5/18 &  2,033/1,300 & 2,281/1,503  &  225/225 & 233/241 &   &  &  \\
D & 10/37 &   2537/700 & 2,816/879  &  281/281 &  322/365 &   &  &  \\

\hline
\end{tabular}
\vspace{-0.1cm}
\caption{Statistics for Positive/Negative Instances in Training, Dev, and Test Sets for Each Experiment.}
\vspace{-0.1cm}
\label{aceStatistics}
\end{table*}

\begin{table*}[!htp]
\center
\footnotesize
\begin{tabular}{p{1.0cm}<{\centering}|p{2.4cm}|p{1.2cm}<{\centering}|p{1.2cm}<{\centering}|p{1.2cm}<{\centering}|p{1.3cm}<{\centering}|p{1.3cm}<{\centering}|p{1.3cm}<{\centering}}
\hline 
\multirow{2}{2.4cm}{Setting} & \multirow{2}{2.4cm}{Method} & \multicolumn{3}{p{4.8cm}<{\centering}|}{Hit@k Trigger Classification (\%)} & \multicolumn{3}{p{4.8cm}<{\centering}}{Hit@k Argument Classification (\%)} \\
\cline{3-8}
& & 1 & 3 & 5 & 1 & 3 & 5 \\
 \hline
& WSD-Embedding    & 1.73  &  13.01  &   22.84  & 2.39 & 2.84 & 2.84      \\
\hline
A & \multirow{4}{2.4cm}{Our Approach}&  3.98 &  23.77  & 32.54 & 1.25 & 3.41 & 3.64  \\
B & & 7.04  & 12.48 & 36.79 & 3.53 & 6.03 & 6.26 \\
C & &  20.05 & 34.66 & 46.48 & 9.56 & 14.68 & 15.70 \\
D & &  33.47 & 51.40 & 68.26 & 14.68 & 26.51 & 27.65     \\
\hline
\end{tabular}
\vspace{-0.1cm}
\caption{Hit@K Performance on Trigger and Argument Classification.}
\vspace{-0.4cm}
\label{acePerformance}
\end{table*}

To show the effectiveness of structural similarity in our approach, we design a baseline, WSD-Embedding, which directly grounds event mentions and arguments to their candidate types and roles using our pre-trained word sense embeddings. Table~\ref{acePerformance} shows that the structural similarity is much more effective than lexical similarity for both trigger and argument classification. Also, as the number of seen types in training increases, the transfer model's performance improves.

We further evaluate the performance of our transfer approach on similar and distinct unseen types. The 33 sub-types defined in ACE fall within 8 coarse-grained main types, such as \textit{Life}, \textit{Justice}. Each subtype belongs to one main type. Subtypes that belong to the same main type tend to have similar structures. For example, \textit{Trial-Hearing} and \textit{Charge-Indict} have the same set of argument roles. 
For training our transfer model, we select 4 subtypes of \textit{Justice}: \textbf{Arrest-Jail}, \textbf{Convict}, \textbf{Charge-Indict}, \textbf{Execute}. For testing, we select another 3 subtypes of \textit{Justice}: \textit{Sentence, Appeal, Release-Parole}. Additionally, we also select one subtype from each of the other seven main types for comparison. %, and test on various other types.
Table~\ref{singleType} shows that, when testing on a new unseen type, the more similar it is to the seen types, the better performance is achieved.
\begin{table}[!htp]
\footnotesize
\begin{tabular}{p{1.1cm}p{2.06cm}|p{0.9cm}<{\centering}|p{0.9cm}<{\centering}|p{0.9cm}<{\centering}}
\hline
\multirow{2}{1.1cm}{Type} & \multirow{2}{1.9cm}{Subtype} & \multicolumn{3}{p{3.67cm}<{\centering}}{Hit@k Trigger Classification} \\
\cline{3-5}
 & & 1 & 3 & 5 \\
\hline
Justice & Sentence & 68.29 & 68.29 & 69.51 \\
Justice & Appeal & 67.50 & 97.50 & 97.50 \\
Justice & Release-Parole & 73.91 & 73.91 & 73.91 \\
\hline
Conflict & Attack & 26.47 & 44.52 & 46.69 \\
Transaction & Transfer-Money & 48.36 & 68.85 & 79.51 \\
Business & Start-Org & 0 & 33.33 & 66.67 \\
Movement & Transport & 2.60 & 3.71 & 7.81 \\
Personnel & End-Position & 9.09 & 50.41 & 53.72 \\
Contact & Phone-Write & 60.78 & 88.24 & 90.20 \\
\hline
Life & Injure & 87.64 & 91.01 & 91.01 \\
\hline
\end{tabular}
\vspace{-0.2cm}
\caption{Performance on Various Types Using Justice Subtypes for Training}
\label{singleType}
\vspace{-0.4cm}
\end{table}

\subsection{ACE Event Identification \& Classification}
\label{subsection:wholeACE}
%\heng{%-    Why selecting 33 ACE event types in the new ontology?
%-    The authors mention the sampling of 20\% of instances and “some”
%sentences without any events annotated: add more details on how these instances
%were selected.}

%Our third experiment compares the performance of our transfer approach to that of two state-of-the-art supervised extraction systems, for which we evaluate their precision, recall, and f-measure on trigger identification and classification as well as on argument identification and classification. 
%Our third experiment compares the performance of our transfer approach to two state-of-the-art supervised extraction methods on both of trigger and argument identification and classification. 
%In this experiment we evaluate the performance of our transfer approach on both of trigger and argument identification and classification. 

Considering that ACE05 corpus includes the richest annotations for event extraction to date, to assess our transferable neural architecture on a large number of unseen types when trained on limited annotations of seen types, we construct a new event ontology which combines 33 ACE event types and argument roles, and 1,161 frames from FrameNet except for the most generic frames such as \textit{Entity}, \textit{Locale}. Some ACE event types easily align to frames, e.g., \textit{Die} is aligned with \textit{Death}%, \textit{Be-Born} is aligned with \textit{Being-Born} in FrameNet
. Some frames are instead more accurately treated as inheritors of ACE types, such as \textit{Suicide-Attack}, which inherits from \textit{Attack}. We manually mapped the selected frames to ACE types.

We compare our approach against the following supervised methods:

\begin{enumerate}[label=$\bullet$]\itemsep0pt
\fontsize{10.6pt}{13pt}\selectfont
\item LSTM: A long short-term memory neural network~\cite{hochreiter1997long} based on distributed semantic features, similar to~\cite{feng2016language}.
\item Joint: A structured perceptron model based on symbolic semantic features~\cite{li2013}. 
\end{enumerate}

For our approach, we follow the experiment setting D in Section~\ref{subsec:ace}, 
%using the same training and development data sets for the 10 seen types,
but target at the 1191 event types in our new event ontology. %, instead of just the 33 ACE event types. 
For evaluation, we sample 150 sentences from the remaining ACE05 data, which contain 129 annotated event mentions for the 23 testing types. % with 91 sentences and sample some sentences without any events annotated. There are 150 sentences in total for evaluation. 
For both LSTM and Joint approaches, we use the entire ACE05 annotated data for 33 ACE event types for training except for the held-out 150 evaluation sentences. %\heng{maybe generate learning curves for all types for the supervised model?, and move section 3.5 here}

We identify the candidate triggers and arguments based on the approach in Section~\ref{subsec:identification}, and map each candidate trigger and argument to the target event ontology. We evaluate on the event mentions which are classified into the 23 testing ACE types. Table~\ref{aceExtractionPerformance} shows the performances. %We can see that, 
%Compared with these two supervised approaches which are trained from 3,464 sentences, though targeting at about 1,194 types, our transfer learning framework can obtain comparable performance on both trigger and argument typing\heng{remove this sentence, and replace it with other analysis on saving annotation cost, if the learning curve can be done}. 

%\heng{change the caption Trigger Typing -> Trigger Identification + Classification; Arg Typing -> Arg Identification+Classification}

%\heng{what is setting D? avoid using weird abbreviation}

\begin{table*}[h]
%\small
\footnotesize
\begin{tabular}{p{0.65cm}<{\centering}|p{1.0cm}<{\centering}|p{0.5cm}<{\centering}p{0.6cm}<{\centering}p{0.6cm}<{\centering}|p{0.6cm}<{\centering}p{0.5cm}<{\centering}p{0.5cm}<{\centering}|p{0.5cm}<{\centering}p{0.6cm}<{\centering}p{0.6cm}<{\centering}|p{0.6cm}<{\centering}p{0.6cm}<{\centering}p{0.6cm}<{\centering}}
\hline 
\multirow{2}{0.65cm}{Setting} & \multirow{2}{1.0cm}{Method} & \multicolumn{3}{p{2.7cm}<{\centering}|}{Trigger Identification} & \multicolumn{3}{p{2.9cm}<{\centering}|}{Trigger Identification + Classification} & \multicolumn{3}{p{2.8cm}<{\centering}|}{Arg Identification} &  \multicolumn{3}{p{2.8cm}<{\centering}}{Arg Identification + Classification} \\
\cline{3-14}
 & & P & R & F & P & R & F & P & R & F & P & R & F \\
 \hline
% \hline
%D &  LSTM     &  59.3 & 54.3  & 56.7 & 55.1 & 50.4 & 52.6 & 47.8 & 22.6 & 30.6 & 28.9 & 13.7 & 18.6 \\
D &  LSTM     &  94.7 & 41.8  & 58.0 & 89.4 & 39.5 & 54.8 & 47.8 & 22.6 & 30.6 & 28.9 & 13.7 & 18.6 \\
D &  Joint    &   55.8 &  67.4 &  61.1   &  50.6 &  61.2  &  55.4  & 36.4 & 28.1 & 31.7 & 33.3 & 25.7 & 29.0  \\
\hline
D &  Transfer & 85.7 & 41.2 & 55.6 & 75.5 & 36.3 & 49.1 &  28.2 & 27.3 & 27.8 & 16.1 & 15.6 & 15.8  \\
\hline
\end{tabular}
\caption{Performance of Trigger and Argument Extraction on ACE Types. (\%)}
\label{aceExtractionPerformance}
\vspace{-0.1cm}
\end{table*}

\begin{figure}[!htb]
\centering
\includegraphics[width=.37\textwidth]{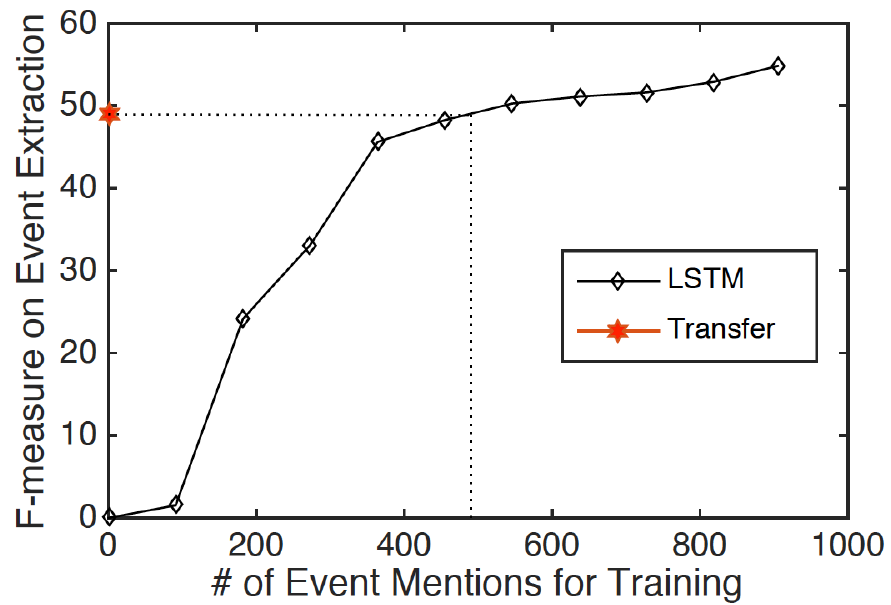}
\vspace{-0.1cm}
\caption{Comparison between Our Approach with LSTM on 23 Testing Event Types.}
\label{curve}
\vspace{-0.3cm}
\end{figure}

%without using any annotations for the testing types, our approach can also achieve a fairly good performance. 
To further demonstrate the zero-shot learning ability of our framework and the significance on saving human annotation effort, we use the supervised LSTM approach for comparison, because it achieved state-of-the-art performance on ACE event extraction~\cite{feng2016language}. The training data of LSTM contains 3,464 sentences with 905 annotated event mentions for the 23 testing event types. We divide these event annotations into 10-fold and successively add another 10\% into the training data of LSTM. Figure~\ref{curve} shows the learning curve. Without any annotated mentions of the 23 test event types in its training set, our transfer learning approach achieves performance comparable to that of the LSTM, which is trained on 3,000 sentences with 500 annotated event mentions. 

In analyzing the triggers which are annotated with ACE types but misclassified into incorrect types or frames, we observe that most errors occur among the types that are defined under the same scenario. 
For example, in the following sentence ``\textit{Abby was a true water \textbf{birth} ( 3kg - normal ) and with Fiona I was dragged out of the pool after the head crowned}'', \textbf{birth} should be a \textit{Being-Born} event while our approach misclassified it as \textit{Giving-Birth} because both \textit{Being-Born} and \textit{Giving-Birth} are defined for \textit{Birth-Scenario} and have very similar predefined roles. 
For argument classification, our approach heavily relies on the semantics of argument path and argument concepts, while many argument roles such as \textit{Entity}, \textit{Organization}, are not informative enough to be matched with argument concepts.

%\subsection{New Type Identification \& Classification}
\subsection{Event Extraction on New Types}
\label{subsec:liberal}
 
In Section~\ref{subsection:wholeACE}, as we use 1,194 event types as the target ontology, we further evaluate the performance of our approach on non-ACE types. From the testing results of Section~\ref{subsection:wholeACE}, we randomly sample 200 event mentions assigned with non-ACE types and ask a linguistic expert to manually assess them. For each mention, the annotator can see its trigger, arguments, the source sentence, the frame and roles assigned by our approach, as well as the definition and examples of the frame from FrameNet\footnote{https://framenet.icsi.berkeley.edu/fndrupal/}. The annotator marks true or false by judging whether the type and argument roles are correct. Table~\ref{allEventTypes} shows the performance.   

\begin{table}[h]
\centering
\small
\begin{tabular}{p{4.8cm}<{\centering}p{2.0cm}<{\centering}}
\hline 
Performance & Accuracy  (\%)\\
\hline
\hline
Trigger Identification + Classification &  40.9 \\
Arg Identification + Classification &  17.41 \\
\hline
\end{tabular}
\vspace{-0.1cm}
\caption{Overall Performance on All Event Types}
\label{allEventTypes}
\vspace{-0.3cm}
\end{table}

Our approach can discover a lot of new events that are not annotated in ACE. For example, in the sentence ``\textit{15 dead as suicide bomber \textbf{blasts} student bus, Israel hits back in Gaza}'', \textbf{blasts} is correctly identified as an \textit{Explosion} event. However, many triggers are mapped to the correct scenario but assigned with incorrect types. For example, in the sentence ``\textit{But Anwar's lawyers said they were \textbf{filing} a fresh request for bail pending a further appeal.}'', \textbf{filing} is identified as a trigger and mapped to the correct \textit{Bail} related scenario but misclassified as a \textit{Bail-Decision} event. We find that, to determine the type of an event mention, besides the consistent semantics between event mentions and type structures, the trigger sense should also be consistent with the definition of the event type. 

\subsection{Impact of AMR}

In our work, we use AMR parsing output to construct event structures. To assess the impact of AMR parser~\cite{wang2015-amr} on event extraction, we choose a subset of ERE corpus which has perfect AMR annotations\footnote{This subset contains 304 documents with 1,022 annotated event mentions of 40 types.}. We select the top-6 most popular event types (\textit{Arrest-Jail, Execute, Die, Meet, Sentence, Charge-Indict}) with 548 manual annotations as seen types. We sample 500 negative event mentions from distinct types of clusters generated from the system~\cite{huangliberal} based on ERE training sentences. We combine the annotated events for seen types and the negative event mentions, and use 90\% for training and 10\% for development. For evaluation, we select 200 sentences from the remaining ERE subset, which contains 128 \textit{Attack} event mentions and 40 \textit{Convict} event mentions. Table~\ref{impactAMR} shows the event extraction performances based on perfect AMR and system AMR respectively. 

Using the same data sets, we further evaluate the performance of our approach using different semantic parsing outputs. We compare AMR with Semantic Role Labeling (SRL) output~\cite{palmer2010semantic} by keeping only the core roles (e.g., \textit{:ARG0, :ARG1}) from AMR annotations. As Table~\ref{impactAMR} shows, compared with SRL output, the fine-grained AMR semantic relations such as \textit{:location}, \textit{:instrument} appear to be more informative to infer the argument roles.

\begin{table}[h]
\centering
\footnotesize
\begin{tabular}{p{2.3cm}|p{0.4cm}<{\centering}|p{0.4cm}<{\centering}|p{0.4cm}<{\centering}|p{0.4cm}<{\centering}|p{0.4cm}<{\centering}|p{0.4cm}<{\centering}}
\hline 
\multirow{2}{2.3cm}{Method} & \multicolumn{3}{p{1.5cm}<{\centering}|}{Trigger $F_{1}$} & \multicolumn{3}{p{1.5cm}<{\centering}}{Arg $F_{1}$} \\
\cline{2-7}
 & P & R & $F_{1}$ & P & R & $F_{1}$ \\ 
 \hline
Perfect AMR & 79.1 & 47.1 & 59.1 & 25.4 & 21.4 & 23.2 \\
Perfect AMR Core Roles (SRL) & 77.1 & 47.0 & 58.4 & 19.7 & 16.9 & 18.2 \\
System AMR & 85.7 & 32.0 & 46.7 & 22.6 & 15.8 & 18.6 \\
\hline
\end{tabular}
\vspace{-0.2cm}
\caption{Impact of AMR Parser and Semantic Information on Trigger and Argument Identification and Classification (\%).}
\label{impactAMR}
\vspace{-0.3cm}
\end{table}

\section{Related Work}
\label{sec:rel}

Most of previous event extraction methods were based on supervised learning using symbolic features~\cite{Ji08,Miwa09,Liao10,Liu10,hong2011using,mcclosky2011event,RiedelM11,chen2012joint,li2013,liuleveraging} or distributional features~\cite{Chen2015,nguyen2015event,feng2016language,nguyen2016joint} from a large amount of training data, regarding event types and argument roles as symbols. Such work can achieve high quality for given types, but cannot be applied to new types without annotation. In contrast, we provide a new angle to vision event extraction and model it as a ``grounding'' task by taking advantage of rich semantics of event types.

Some other IE paradigms such as Open IE ~\cite{Etzioni05unsupervisednamed-entity,banko2007open,banko2008tradeoffs,etzioni2011open,ritter2012open}, Pre-emptive IE~\cite{shinyama2006preemptive}, On-demand IE~\cite{sekine2006demand}, Liberal IE~\cite{huangliberal,huang2017liberal}, and semantic frame based event discovery~\cite{Kim2013} can discover many events without pre-defined event schema. The event types and argument roles are inferred by a cluster of similar events. These paradigms heavily rely on information redundancy, so cannot work when the input consists of only a few sentences. Our work can discover events from any size of given corpus and can also be complementary with these paradigms because it can ground each event cluster to a rich predefined event ontology. 

Zero-Shot learning has been widely applied in visual object classification~\cite{frome2013devise,norouzi2013zero,socher2013zero}, fine-grained name tagging~\cite{malabel,qu2016named} and relation extraction~\cite{verga2015multilingual}. Different from these tasks, the seen types in event extraction are limited. The most popular event schemas, such as ACE, only defined 33 event types while most visual object training sets contain more than 1,000 types. Thus, the methods proposed for zero-shot visual object classification cannot be directly applied to event extraction because of overfitting. Thus, we design a new loss function by creating  ``negative'' training instances to avoid overfitting. 

\section{Conclusions and Future Work}
\label{sec:conclusion}

In this work, we take a fresh look at the event extraction task and model it as a grounding problem. We propose a transferable neural architecture, which leverages existing human constructed event schemas and manual annotations for a small set of seen types, and transfers the knowledge from the existing types to the extraction of unseen types, to improve the scalability of event extraction as well as save human effort. Without any annotation, our approach can achieve comparable performance with state-of-the-art supervised models trained from a large amount of labeled data. In the future, we will extend our framework by incorporating event definitions and argument descriptions to improve the event extraction performance.

%\heng{citations are still not consistent, some have full author names and some not, some venue names have years some not, fix them}
% \section*{Acknowledgments}

% Do not number the acknowledgment section.

\bibliography{emnlp2017}
\bibliographystyle{emnlp_natbib}

\end{document}